\pdfoutput=1
\documentclass[11pt]{article}

\usepackage[]{acl}

\usepackage{times}
\usepackage{latexsym}
\usepackage{color}
\usepackage{multirow}
\usepackage{booktabs}
\usepackage{amsmath}

\usepackage{color}
\usepackage{booktabs}
\usepackage{arydshln}
\usepackage[T1]{fontenc}
\usepackage[utf8]{inputenc}

\usepackage{microtype}

\usepackage{inconsolata}

\usepackage{graphicx}

\title{Revisiting Overthinking \protect\\ in Long Chain-of-Thought from the Perspective of Self-Doubt}

\author{%
  Keqin Peng$^{1}$,
  Liang Ding$^{2}$\thanks{~~Corresponding Authors.},
  Yuanxin Ouyang$^{1}$,
  Meng Fang$^{3}$,
  Dacheng Tao$^{4}$\\
  $^{1}$Beihang University $^{2}$The University of Sydney\\ $^{3}$University of Liverpool $^{4}$Nanyang Technological University\\
  \texttt{keqin.peng@buaa.edu.cn},
  \texttt{liangding.liam@gmail.com}}

\begin{document}
\maketitle
\begin{abstract}
% overthinking 的问题很严重
% 对overthinking的问题进行了研究，但是都是定性的研究，因此我们定量的分析self-doubt的影响，并提出一个简单的有效的方法提升模型效果
Reasoning Large Language Models (RLLMs) have demonstrated impressive performance on complex tasks, largely due to the adoption of Long Chain-of-Thought (Long CoT) reasoning. However, they often exhibit overthinking --- performing unnecessary reasoning steps even after arriving at the correct answer. Prior work has largely focused on qualitative analyses of overthinking through sample-based observations of long CoTs. In contrast, we present a quantitative analysis of overthinking from the perspective of self-doubt, characterized by excessive token usage devoted to re-verifying already-correct answer. We find that self-doubt significantly contributes to overthinking. In response, we introduce a simple and effective prompting method to reduce the model's over-reliance on input questions, thereby avoiding self-doubt. Specifically, we first prompt the model to question the validity of the input question, and then respond concisely based on the outcome of that evaluation. Experiments on three mathematical reasoning tasks and four datasets with missing premises demonstrate that our method substantially reduces answer length and yields significant improvements across nearly all datasets upon 4 widely-used RLLMs. Further analysis demonstrates that our method effectively minimizes the number of reasoning steps and reduces self-doubt.

\end{abstract}

\section{Introduction}

\begin{figure*}[t]
\centering
  \includegraphics[width=1.0\linewidth]{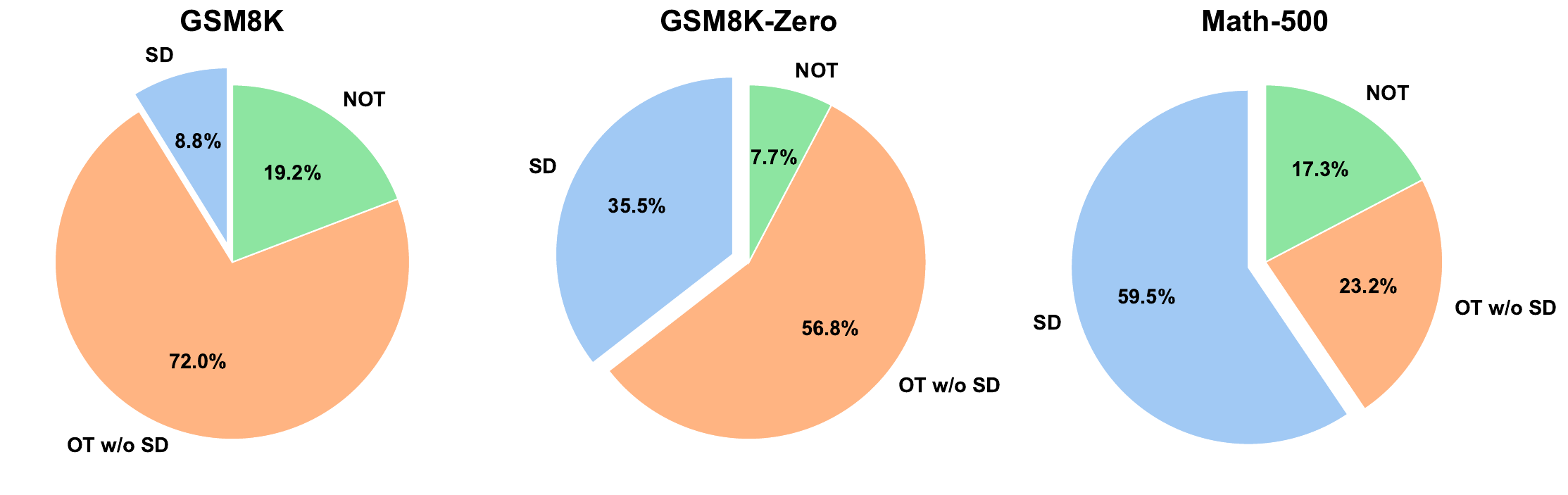}
  \caption{\textbf{The proportion of overthinking and self-doubt across three different datasets with \textit{Deepseek-Deepseek-R1-Distill-Qwen-32B}.} SD: Self-Doubt, OT w/o SD: Overthinking without Self-Doubt, NOT: Non-Overtinking.}
  \label{fig:sd_32B}
\end{figure*}

Recent advances in reasoning models, such as OpenAI’s o1~\cite{jaech2024openai} and DeepSeekR1~\cite{guo2025deepseek}, have demonstrated significant progress in complex reasoning capabilities, particularly in domains such as mathematical reasoning~\cite{wei2022chain}. A central factor in their success lies in the application of long chain-of-thought (Long CoT), a kind of in-context learning~\cite{dong-etal-2024-survey,peng-etal-2024-revisiting}, which enhances reasoning abilities by a more detailed, iterative process of exploration and reflection~\cite{chen2025towards}.

However, while long CoT reasoning significantly boosts accuracy, step-by-step thinking mechanisms also lead to lengthy output responses, resulting in substantial computational overhead and increased reasoning time~\cite{chen2024not,sui2025stop}. Previous studies have predominantly employed qualitative analyses of the overthinking phenomenon by examining the observed reasoning paths. For example, ~\citet{chen2024not} found that Reasoning models will generate hundreds of response tokens even for simple questions like \textit{"2+3=?"}; ~\citet{fu2025reasoning} identify self-doubt, which repeats check and verify when obtaining the correct information, as one major source of overthinking; ~\citet{fan2025missing} found that an ill-posed question with missing premises will exacerbate overthinking. To conduct a more in-depth analysis of the phenomenon of overthinking, we performed a quantitative analysis of overthinking from the perspective of self-doubt. Specifically, we categorize the generated reasoning paths into three classes: overthinking with Self-Doubt (SD), overthinking without Self-Doubt (OT w/o SD), and non-overthinking (NOT). We then employ a third-party LLM to evaluate these reasoning paths. Experimental results reveal that overthinking is widespread, and self-doubt significantly contributes to overthinking, especially in complex tasks.

Inspired by the social comparison theory~\cite{festinger1957social}, we attribute the self-doubt observed in long CoT to the model's excessive deference to the user, manifested as an overreliance on user input and a strong desire to please, which makes it overly cautious and reluctant to challenge user inputs.
In response, we introduce a simple and effective prompting method to mitigate the overthinking phenomenon in Long CoT. 
Specifically, we first prompt the model to assess the validity of the input question, thereby engaging its critique abilities and boosting its confidence; We then instruct the model to provide a concise response based on the outcome of this evaluation. 
Experiments on three mathematical reasoning tasks demonstrate that our method substantially reduces answer length and yields significant improvements across nearly all datasets upon 4 widely-used RLLMs. Further experiments on four datasets with missing premises demonstrate that our method significantly reduces token consumption and more effectively identifies cases where premises are missing.

\section{Revisiting Overthinking}
\label{sec:revisit}

Recently, overthinking in the Long Chain-of-Thought has received widespread attention due to the huge success of RLLM. Many researchers~\cite{chen2025towards} have explored the causes and characteristics of overthinking through qualitative analysis of reasoning paths. In this section, we adopt the lens of self-doubt to quantitatively investigate its influence on the overthinking phenomenon. To systematically assess the impact of self-doubt on overthinking, we conduct experiments across diverse datasets using a widely used reasoning model. Specifically, we generate the reasoning path for three mathematical reasoning tasks, GSM8K~\cite{cobbe2021training}, GSM8K-Zero~\cite{chiang2024over}, and Math-500~\cite{lightman2023let}, with reasoning model \textit{Deepseek-R1-Distill-Qwen-32B}~\cite{guo2025deepseek}. For evaluation, we categorize each reasoning path into one of three groups:

\begin{itemize}
\item \textbf{Overthinking with Self-Doubt (SD)}: The model's answer includes unnecessarily long reasoning and repeatedly verifies information that is already correct.
\item \textbf{Overthinking without Self-Doubt (OT w/o SD)}: The answer is excessively long but does not involve redundant verification.
\item \textbf{Non-Overthinking (NOT)}: The answer avoids both unnecessary reasoning and redundant verification.
\end{itemize}

We employ Qwen2.5-72B-Instruct~\cite{qwen2.5} as an automated evaluator. The full evaluation prompt is provided in Appendix~\ref{sec: eval}. Figure~\ref{fig:sd_32B} reveals that the overthinking phenomenon is prevalent in Long CoT reasoning. Specifically, the proportion of overthinking samples (comprising both self-doubt and overthinking, excluding self-doubt) exceeds 80\% across all three datasets, even reaching as high as 92\% in the GSM8K-Zero dataset. Furthermore, the figure highlights that self-doubt is a major contributor to overthinking, particularly in the complex math-500 task, where it accounts for nearly 60\% of cases, underscoring the importance of self-doubt in the overthinking process.

\begin{table*}
\centering
\resizebox{1.0\linewidth}{!}{
\begin{tabular}{cccccccccc} 
\toprule
\multirow{2}{*}{\textbf{Model}}          & \multirow{2}{*}{\textbf{Method}} & \multicolumn{2}{c}{\textbf{GSM8K}}                             & \multicolumn{2}{c}{\textbf{GSM8K-Zero}}                        & \multicolumn{2}{c}{\textbf{MATH-500}}                           & \multicolumn{2}{c}{\textit{\textbf{Average}}}                    \\ 
\cmidrule(lr){3-4}\cmidrule(r){5-6}\cmidrule(r){7-8}\cmidrule(lr){9-9}\cmidrule(lr){10-10}
                                         &                                  & \textbf{Length}               & \textbf{Accuracy}              & \textbf{Length}               & \textbf{Accuracy}              & \textbf{Length}                & \textbf{Accuracy}              & \textbf{Length}                & \textbf{Accuracy}               \\ 
\midrule
\multirow{2}{*}{\textbf{DS Distill 14B}} & \textbf{baseline}                & 591                           & 92.4                           & 1070                          & 80.6                           & 3095                           & \textbf{\textcolor{red}{92.8}} & 1585                           & 88.6                            \\
                                         & \textbf{Ours}                    & \textcolor{red}{\textbf{426}} & \textbf{\textcolor{red}{92.4}} & \textcolor{red}{\textbf{392}} & \textbf{\textcolor{red}{86.2}} & \textcolor{red}{\textbf{1922}} & 92.2                           & \textcolor{red}{\textbf{913}}  & \textbf{\textcolor{red}{90.3}}  \\ 
\midrule
\multirow{2}{*}{\textbf{DS Distill 32B}} & \textbf{baseline}                & \textbf{\textcolor{red}{551}} & 94.2                           & 613                           & 72.9                           & 3058                           & 92.4                           & 1407                           & 86.5                            \\
                                         & \textbf{Ours}                    & 562                           & \textbf{\textcolor{red}{95.2}} & \textbf{\textcolor{red}{495}} & \textbf{\textcolor{red}{89.6}} & \textcolor{red}{\textbf{2141}} & \textbf{\textcolor{red}{93.6}} & \textcolor{red}{\textbf{1066}} & \textbf{\textcolor{red}{92.8}}  \\ 
\midrule
\multirow{2}{*}{\textbf{DS Distill 70B}} & \textbf{baseline}                & 555                           & 94.2                           & 604                           & 74.8                           & 2847                           & \textbf{\textcolor{red}{92.4}} & 1335                           & 87.1                            \\
                                         & \textbf{Ours}                    & \textcolor{red}{\textbf{434}} & \textcolor{red}{\textbf{95.1}} & \textcolor{red}{\textbf{438}} & \textcolor{red}{\textbf{89.5}} & \textcolor{red}{\textbf{1927}} & 91.6                           & \textcolor{red}{\textbf{933}}  & \textcolor{red}{\textbf{92.1}}  \\ 
\midrule
\multirow{2}{*}{\textbf{Qwen3-32B}}      & \textbf{baseline}                & 1681                          & \textbf{\textcolor{red}{95.8}} & 1854                          & 91.0                           & 3970                           & \textbf{\textcolor{red}{97.2}} & 2502                           & \textbf{\textcolor{red}{94.7}}  \\
                                         & \textbf{Ours}                    & \textcolor{red}{\textbf{631}} & 94.6                           & \textcolor{red}{\textbf{542}} & \textcolor{red}{\textbf{92.0}} & \textcolor{red}{\textbf{2981}} & 96.8                           & \textcolor{red}{\textbf{1385}} & 94.5                            \\
\bottomrule
\end{tabular}}
\caption{\textbf{Performance of our method across three well-defined models} on four reasoning models. The best results are in \textcolor{red}{red}. On average, we reduce \textit{reasoning length} by \textbf{-37.1\%} while improving \textit{accuracy} by \textbf{+3.6\%}.}
\label{tab: main_normal}
\end{table*}

\section{Method}
Social comparison theory~\cite{festinger1957social} holds that when someone relies heavily on others’ opinions or approval to judge themselves, it can lead to self-doubt if those comparisons are unfavorable or inconsistent. Similarly, large language models, due to their heavy dependence on human evaluations, continuously engage in self-doubt and self-verification to produce answers that satisfy users, which can lead to the phenomenon of overthinking. 
To mitigate this, we propose a simple yet effective prompting strategy to elicit the model's critical ability, thereby alleviating overthinking. Specifically, we first prompt the model to assess the validity of the input query. Based on the evaluation outcome, we then instruct the model to provide a concise answer or indicate what information is missing. The full templates of prompts are presented in Table~\ref{tab:prompts}.

\begin{table}[]
\setlength{\tabcolsep}{4pt}
\centering
\resizebox{1\columnwidth}{!}{
\begin{tabular}{l p{7cm}}
\toprule
\bf Method & \multicolumn{1}{c}{\bf Prompt Templates} \\
\midrule
\textbf{Baseline} & \texttt{"role": "user", "content": [Question]} \\
\hdashline
\textbf{Ours} & \texttt{"role": "user", "content": [Question] Before reasoning deeply, check whether all necessary information is available. If any key data is missing or ambiguous, explicitly state that first; otherwise, answer it with the minimum number of tokens required.} \\
\bottomrule
\end{tabular}}
\caption{\textbf{The prompt template} of our method.}
\label{tab:prompts}
\end{table}

\section{Experimental Setup}
\paragraph{Models and Baselines.} We perform experiments across different sizes of reasoning models, including \textit{DeepSeek-R1-Distill-Qwen-14B}, \textit{DeepSeek-R1-Distill-Qwen-32B}, \textit{DeepSeek-R1-Distill-Llama-70B}~\cite{guo2025deepseek} and \textit{Qwen3-32B}~\cite{yang2025qwen3}, which are all widely-used RLLMs. For the baseline, we utilize the default chat template without any additional information. We set the temperature to 0 by default to obtain deterministic output, as proposed in ~\citet{peng-etal-2023-towards}.

\paragraph{Datasets.} We conduct a systematic study across three mathematical reasoning tasks, including GSM8K~\cite{cobbe2021training}, GSM8K-Zero~\cite{chiang2024over}, and Math-500~\cite{lightman2023let}. And GSM8K-Zero, whose answer can be easily generated by LLMs, specifically targets the analysis of over-reasoning and redundancy in LLM-generated outputs. We also evaluate our method in ill-posed questions with missing premises (MiP), which will exacerbate Overthinking~\cite{fan2025missing}, and we follow the setting of ~\citet{fan2025missing} using four MiP datasets: MiP-Formula, MiP-SVAMP, MiP-GSM8K, and MiP-Math. 
 
\paragraph{Evaluation Metrics.} To evaluate the answer quality in different datasets, we follow the setting of ~\citet{fan2025missing} to adopt both statistic metric and LLM-based metric~\cite{zheng2023judging}, \textit{i.e.}, \textbf{LLM-as-a-Judge}. Specifically, for MiP datasets, we use \textbf{Abstain Rate}, defined as the proportion of answers in which the model successfully identifies a missing premise. For the well-defined questions, we employ \textbf{Accuracy}, the proportion of answers where the model produces a definitive response that aligns with the reference answer. Additionally, we report the \textbf{Response Length}, capturing both the reasoning process and the final answer, to assess tendencies toward overthinking. Evaluation protocols are provided in Appendix~\ref{sec: eval}.

\begin{table*}
\centering
\resizebox{1.0\linewidth}{!}{
\begin{tabular}{cccccccccc} 
\toprule
\multirow{2}{*}{\textbf{Model}}                  & \multirow{2}{*}{\textbf{Method}} & \multicolumn{2}{c}{\textbf{MiP-Formula}}                        & \multicolumn{2}{c}{\textbf{MiP-SVAMP}}                         & \multicolumn{2}{c}{\textbf{MiP-GSM8K}}                                   & \multicolumn{2}{c}{\textbf{\textbf{MiP-MATH}}}  \\ 
\cmidrule(r){3-4}\cmidrule(lr){5-6}\cmidrule(lr){7-8}\cmidrule(r){9-10}
                                                 &                                  & \textbf{Length}                & \textbf{Abstain}              & \textbf{Length}               & \textbf{Abstain}              & \textbf{Length}               & \textbf{Abstain}              & \textbf{Length}                & \textbf{Abstain}                               \\ 
\midrule
\multirow{2}{*}{\textbf{DS Distill 14B}} & \textbf{baseline}                & 5740                           & 54.0                           & 1695                          & 76.0                           & 1552                          & 32.5                           & 6106                           & 54.9                                            \\
                                                 & \textbf{Ours}                    & \textcolor{red}{\textbf{735}}  & \textcolor{red}{\textbf{96.0}} & \textcolor{red}{\textbf{413}} & \textcolor{red}{\textbf{96.7}} & \textcolor{red}{\textbf{327}} & \textcolor{red}{\textbf{73.7}} & \textcolor{red}{\textbf{2406}} & \textcolor{red}{\textbf{86.3}}                  \\ 
\midrule
\multirow{2}{*}{\textbf{DS Distill 32B}} & \textbf{baseline}                & 5633                           & 66.0                           & 1060                          & 78.0                           & 1249                          & 38.1                           & 6414                           & 67.3                                            \\
                                                 & \textbf{Ours}                    & \textcolor{red}{\textbf{1060}}  & \textcolor{red}{\textbf{92.0}} & \textcolor{red}{\textbf{451}} & \textcolor{red}{\textbf{97.0}} & \textcolor{red}{\textbf{446}} & \textcolor{red}{\textbf{79.6}} & \textcolor{red}{\textbf{2789}} & \textcolor{red}{\textbf{76.9}}                  \\ 
\midrule
\multirow{2}{*}{\textbf{DS Distill 70B}}   & \textbf{baseline}                & 4761                           & 52.0                           & 812                           & 80.3                           & 973                           & 28.9                           & 6286                           & 61.5                                            \\
                                                 & \textbf{Ours}                    & \textcolor{red}{\textbf{529}}  & \textcolor{red}{\textbf{88.0}} & \textcolor{red}{\textbf{428}} & \textcolor{red}{\textbf{96.3}} & \textcolor{red}{\textbf{415}} & \textcolor{red}{\textbf{78.2}} & \textcolor{red}{\textbf{2728}} & \textcolor{red}{\textbf{68.6}}                  \\ 
\midrule
\multirow{2}{*}{\textbf{Qwen3-32B}}              & \textbf{baseline}                         & 5039                           & 36.0                           & 1771                          & 82.0                           & 2957                          & 43.0                           & 6081                           & 61.5                                            \\
                                                 & \textbf{Ours}                    & \textcolor{red}{\textbf{631}} & \textcolor{red}{\textbf{98.0}} & \textcolor{red}{\textbf{319}} & \textcolor{red}{\textbf{97.0}} & \textcolor{red}{\textbf{530}} & \textcolor{red}{\textbf{81.9}} & \textcolor{red}{\textbf{3669}} & \textcolor{red}{\textbf{82.4}}                  \\
\bottomrule
\end{tabular}}
\caption{\textbf{Performance of our method across four Missing Premise (MiP) datasets} on four reasoning models. The best results are in \textcolor{red}{red}. We can see that our method largely improves the abstain rate and significantly reduces the response length.}
\label{tab: main_mip}
\end{table*}

\section{Main Results}
We demonstrate the effectiveness of our method in 3 well-defined mathematical reasoning tasks and four Missing Premise (MiP) datasets across four RLLMs. We summarize the results in Table~\ref{tab: main_normal} and Table~\ref{tab: main_mip}. Based on the results, we can find that:

\paragraph{Our method significantly reduces token consumption while maintaining accuracy in well-defined mathematical reasoning tasks.} Results on ~\ref{tab: main_normal} show that our method can consistently significantly reduce token consumption across different datasets and different reasoning models. Specifically, our method can reduce token consumption by 1/3 on average compared to the baseline. Especially in the GSM8K-Zero dataset under Qwen3-32B, our method brings 2/3 token consumption reduction, showing that our method can effectively alleviate the overthinking phenomenon. Furthermore, we can notice that our method can maintain the accuracy on almost all datasets and even bring improvements on some datasets. In particular, on GSM8K-Zero, our method brings +16.7\% and +14.7\% improvement with \textit{DeepSeek-R1-Distill-Qwen-32B} and \textit{DeepSeek-R1-Distill-Llama-70B}, respectively. We attribute these gains primarily to the correction of errors caused by overthinking.

\paragraph{Our method brings consistent and significant improvement and token reduction in ill-posed questions with missing premises (MiP).} To address the issue of overthinking caused by missing premises, we evaluate our method on the MiP dataset.  As shown in Table~\ref{tab: main_mip}, our method consistently outperforms the baseline across all datasets in terms of both response length and abstain rate. Specifically, it reduces token consumption by at least 50\% on nearly all datasets. It’s worth highlighting that our method reduces token consumption by more than 80\% on the MiP-Formula dataset, from more than 5,000 tokens to about 1,000. Additionally, our approach significantly improves abstain rates, achieving gains of nearly 40\% on MiP-Formula and MiP-GSM8K. These results highlight our method's effectiveness in reducing overthinking while enhancing the model's critique capability.

\section{Analysis}
\begin{figure}[t]
\centering
  \includegraphics[width=1.0\linewidth]{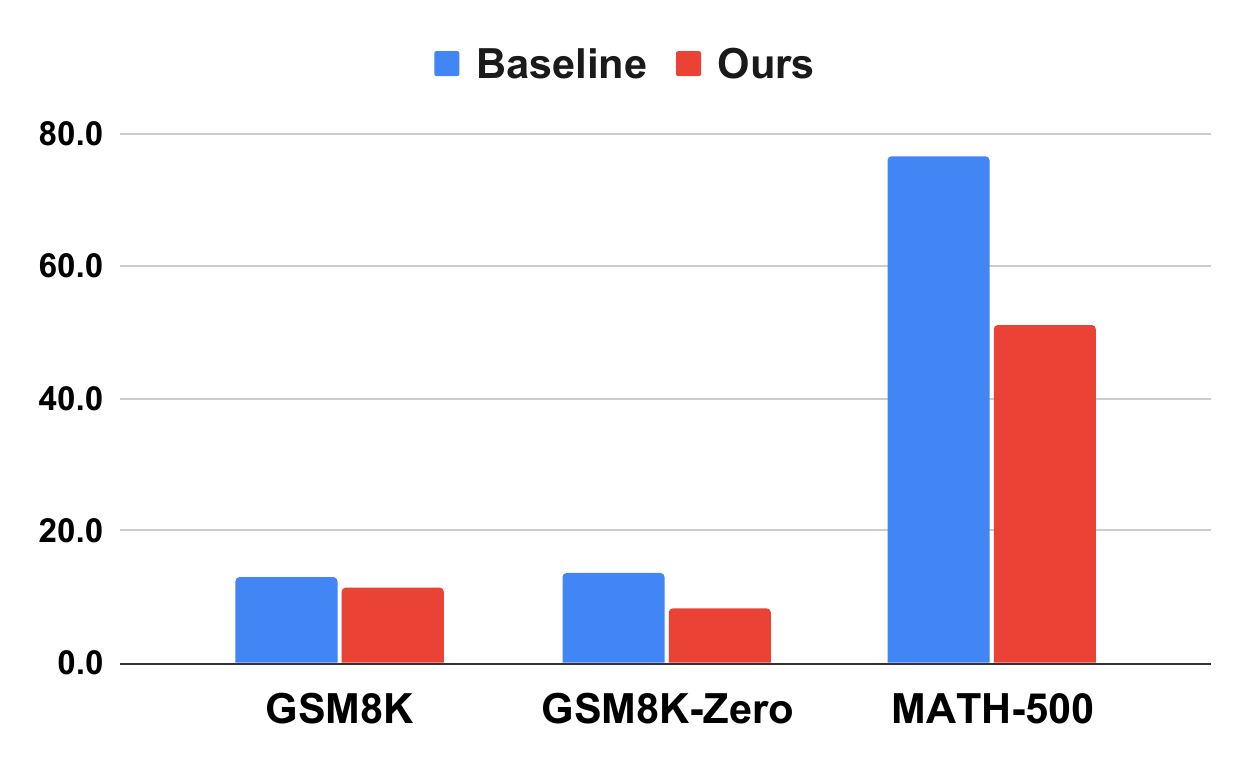}
  \caption{\textbf{The average thinking steps} across three mathematical reasoning tasks with \textit{DeepSeek-R1-Distill-Qwen-32B}.}
  \label{fig:step}
\end{figure}

\paragraph{Our method can significantly reduce the number of reasoning steps in Long Chain-of-Thought.} Self-doubt often causes the model to generate more steps in its thought process. To validate whether our method can reduce self-doubt, we first compare our method with the baseline from thinking patterns. Specifically, we split the generated text with \textit{\textbackslash n\textbackslash n} to obtain the number of reasoning steps. Figure~\ref{fig:step} represents the average thinking steps across three datasets with \textit{DeepSeek-R1-Distill-Qwen-32B}. From Figure~\ref{fig:step}, we can see that our method can consistently reduce the thinking steps, demonstrating that our method can effectively alleviate overthinking by avoiding self-doubt.

\begin{table}
\centering
\resizebox{1.0\columnwidth}{!}{
\begin{tabular}{cccc} 
\toprule
                  & \textbf{GSM8K}                             & \textbf{GSM8K-Zero}            & \textbf{MATH-500}               \\ 
\midrule
\textbf{baseline} & 5.9                                        & 13.5                           & 62                              \\
\textbf{Ours}     & 17.2                                       & 10.1                           & 38.2                            \\ 
\hdashline
\textbf{\textit{$\Delta$}}    & \textbf{\textcolor{green}{-11.3}} & \textcolor{red}{\textbf{+3.5}} & \textbf{\textcolor{red}{23.8}}  \\
\bottomrule
\end{tabular}}
\caption{\textbf{The ratio of self-doubt} across three datasets.}
\label{tab:sd}
\end{table}

\paragraph{Our method can effectively reduce the self-doubt in overthinking, especially in complex datasets.} To verify whether our method effectively reduces self-doubt in overthinking, we report the performance using the evaluation metric described in Section \ref{sec:revisit}. As shown in Table~\ref {tab:sd}, our method reduces the self-doubt ratio in two out of three datasets, with a particularly notable 23.8\% reduction on the complex dataset \textit{MATH-500}, demonstrating our method can effectively reduce self-doubt. Regarding the increase in self-doubt observed in \textit{GSM8K}, we suspect it stems from the early verification of the question's correctness and the short reasoning path. 

\section{Related Works}

\paragraph{Chain-of-Thought.}
Chain‐of‐Thought (CoT) prompting elicits multi‐step reasoning by having LLMs generate intermediate “thought” traces, substantially improving performance on arithmetic, commonsense, and symbolic benchmarks.  Early work showed CoT can unlock latent reasoning capabilities in models like GPT-3 and PaLM~\cite{wei2022chain}. Many follow-up works have shown the effectiveness of carefully-designed CoT prompting on various tasks~\cite{zhong2024achieving,zhang2025intention}, and recent surveys have catalogued extensions to longer, more detailed thought chains~\cite{chen2025towards}.

\paragraph{Overthinking in Long CoT.}
Despite boosted accuracy, long CoTs often induce \emph{overthinking} problem, where models continue producing needless steps or repeat verifications after arriving at the correct answer. \citet{chen2024not,sui2025stop} provide qualitative analyses of these redundant patterns, while \citet{fan2025missing} demonstrate that missing premises exacerbate overthinking in ill‐posed questions. \citet{fu2025reasoning} attribute much of this to \emph{self‐doubt}—models probing their own certainty—and propose certainty‐probing probes to trim reasoning length.

\paragraph{Prompting and Reinforcement for Efficient Reasoning.}
To curb overthinking, several works have explored targeted interventions. \citet{guo2025deepseek} use reinforcement‐learning to incentivize concise reasoning, \citet{zheng2023judging} introduce LLM‐as‐a‐Judge for automated quality assessment, and \citet{fan2025missing} design prompts to detect missing information before reasoning.  Our method builds on these by adding a lightweight input‐validation step that directly mitigates self‐doubt and over‐reliance on user queries.

\section{Conclusion}
To further explore overthinking in reasoning large language models, we conduct a quantitative analysis of overthinking from the perspective of self-doubt. We find that self-doubt is one of the main causes of overthinking. In response, we propose a simple and effective prompt method to reduce self-doubt by eliciting the critique ability. Our experiments across various datasets and models demonstrate that our method can effectively reduce token consumption and improve performance.

\section*{Limitations}
Despite the encouraging results, our study has several important limitations that future work should address: 1) \textbf{Domain and Task Scope}: We evaluate our prompting strategy solely on three mathematical reasoning benchmarks (GSM8K, GSM8K-Zero, Math-500) and four missing-premise (MiP) datasets. It remains unclear whether the same reduction in overthinking and self-doubt would hold for other reasoning domains (e.g., commonsense QA, multi-step planning, or multimodal tasks). 2) \textbf{Reliance on Automated Evaluation}: Our quantitative analysis of overthinking and self-doubt depends on an LLM-based judge (Qwen2.5-72B-Instruct) and a simple paragraph-segmentation heuristic to categorize reasoning paths. This introduces potential biases or misclassifications, since the evaluator itself may misunderstand nuanced reasoning patterns or miss subtle verification steps. Addressing these limitations will be crucial for deploying more efficient and trustworthy chain-of-thought systems in real-world applications.

\bibliography{arxiv_0529}

\newpage
\appendix
% \section{Prompt Template of Our Method}
% \label{sec: app}
% We present the prompt template of our method in Table~\ref{tab:prompts}.

\section{Prompt Template for Evaluation}
\label{sec: eval}
As we need LLM-as-a-judge to evaluate the quality of the generation of the models in various experiments in this study. In this section, we showcase the prompt template we use for each kind of evaluation.

To evaluate self-doubt in long chain-of-thought (CoT) responses, we adopt the template shown in Table~\ref{tab:prompt}. Since the generated long CoTs are often lengthy and self-doubt typically emerges later in the response, we split the output into two segments with the same reasoning steps using the delimiter \textit{\textbackslash n\textbackslash n}. We then use the second segment to assess the proportion of self-doubt in the dataset.

For the evaluation of the models’ answer accuracy and abstain rate, we follow ~\citet{fan2025missing} to adopt the following prompt templates designed for ’paired’ and ’non-paired’ data, respectively. Since RLLMs often output an additional \textit{\textbackslash n\textbackslash n} at the end of response, we also take the last two paragraph segmented by \textit{\textbackslash n\textbackslash n} to avoid passing in an empty string.

\begin{table*}[]
    \centering
    %\vspace{2.8mm}
    \begin{tabular}{p{0.94\linewidth}}
    \toprule
\textit{~Prompt Template for Response Evaluation of MiP Datasets} \\
\vspace{-1mm}
\textbf{System:} You are a helpful assistant that evaluates the quality of a model’s answer. You will be given a question and a model’s answer. You need to evaluate the correctness of the model’s answer. If the answer explicitly says that the condition of the question is insufficient, you should return 0. If the model provides an answer that is a number or formula with variables, you should return 1. Please only return the number, no other text. \\
\textbf{User:} Model answer: [model answer short]\textbackslash n
\\\midrule
\textit{~Prompt Template for Response Evaluation of well-defined Datasets} \\
\vspace{-1mm}
\textbf{System:} You are a helpful assistant that evaluates the quality of a model’s answer. You will be given a question and a model’s answer. You need to evaluate the correctness of the model’s answer. If the model output says that the condition of the question is insufficient, you should return 0. Otherwise, if the model give a clear answer and matches the reference answer, you should return 1. If the model’s answer does not match the reference answer, you should return 2. Please only return the number, no other text. \\
\textbf{User:} Model answer: [model answer]\textbackslash nReference answer: [reference answer]\textbackslash n 
\\\midrule
\textit{~Prompt Template for Self-Doubt Evaluation} \\
\vspace{-1mm}
\textbf{System:} You are a helpful assistant tasked with evaluating the quality of a model’s answer. You will be given a question, a reference answer and the model’s answer. Evaluate the model’s answer according to the following criteria:

- Return 0 if the answer includes unnecessarily long reasoning and repeats verification of already correct answer;

- Return 1 if the answer is unnecessarily long but does not include redundant verification;

- Return 2 if the answer avoids both unnecessary reasoning and redundant verification.

Return only the appropriate number. Do not include any explanation or extra text.

\textbf{User:} Model answer: [model answer]\textbackslash nReference answer: [reference answer]\textbackslash n  
\\\bottomrule
\end{tabular}
\caption{Prompt Template for Response Evaluation.}
\label{tab:prompt}
\end{table*}

\section{Case Study}
We present a case on GSM8K-Zero with \textit{DeepSeek-R1-Distill-Qwen-32B} in Table~\ref{tab:case}, we can notice that baseline method will repeat reflection and our method can largely reduce the answer length.

\begin{table*}[]
    \centering
    %\vspace{2.8mm}
    \begin{tabular}{p{0.94\linewidth}}
    \toprule
\vspace{-1mm}
\textbf{Question:} Weng earns \$12 an hour for babysitting. Yesterday, she just did 50 minutes of babysitting. How many minutes of babysitting did she do yesterday? \\
\\
\textbf{Baseline:} "Okay, so I have this problem here: Weng earns \$12 an hour for babysitting. Yesterday, she just did 50 minutes of babysitting. How many minutes of babysitting did she do yesterday? Hmm, wait a second. That seems a bit confusing. Let me read it again to make sure I understand."\textbackslash n\textbackslash n...That seems straightforward, but maybe I'm missing something. Perhaps it's a trick question or there's more to it....So, maybe it's just a simple question where the answer is 50 minutes....So, I think the answer is 50 minutes.\textbackslash n\textbackslash nBut just to be thorough, let me consider if there's another angle....Wait, perhaps the question is trying to set up a follow-up question, like how much she earned, but the way it's phrased is confusing. The question gives the time she worked and asks for the same information, so the answer is 50 minutes. (\textcolor{red}{476 tokens}) \\
\\
\textbf{Ours:} Okay, so I need to figure out how many minutes Weng did babysitting yesterday. The problem says she earns \$12 an hour and that she just did 50 minutes of babysitting. Wait, that seems a bit confusing. Let me read it again....So, is the answer just 50 minutes? That seems too simple, but maybe that's the case....but the way it's phrased, it's just asking for the minutes she worked, which is given as 50. So, I think the answer is 50 minutes. (\textcolor{red}{304 tokens.})
\\\bottomrule
\end{tabular}
\caption{A case study on GSM8K-Zero with \textit{DeepSeek-R1-Distill-Qwen-32B}. We can see that our method reduce the token consumption.}
\label{tab:case}
\end{table*}

\end{document}